\pdfoutput=1

\documentclass[11pt]{article}

\usepackage[preprint]{acl}

\usepackage{times}
\usepackage{latexsym}
\usepackage{amsmath}
\usepackage{amssymb}
\usepackage{times}
\usepackage{algorithm}
\usepackage{algpseudocode}
\usepackage{mathtools}
\usepackage{fancyvrb}
\usepackage{fancyhdr}
\usepackage{makecell}
\usepackage{booktabs}
\usepackage{amsmath}
\usepackage{listings}
\usepackage[normalem]{ulem}

\usepackage[T1]{fontenc}

\usepackage[utf8]{inputenc}

\usepackage{microtype}

\usepackage{inconsolata}

\usepackage{fancyvrb}

\usepackage{graphicx}
\usepackage{multirow}

\newcommand{\dsetname}{MultiVENT-G}

\title{Grounding Partially-Defined Events in Multimodal Data}

\author{Kate Sanders$^{\spadesuit}$* \hspace{2.5mm} Reno Kriz$^{\spadesuit}$* \hspace{2.5mm} David Etter$^{\spadesuit}$* \hspace{2.5mm} Hannah Recknor$^{\spadesuit}$ \\ \textbf{Alexander Martin}$^{\spadesuit\heartsuit}$ \hspace{2.5mm} \textbf{Cameron Carpenter}$^{\spadesuit}$ \hspace{2.5mm} \textbf{Jingyang Lin}$^{\heartsuit}$ \hspace{2.5mm} \textbf{Benjamin Van Durme}$^{\spadesuit}$ \\
        $^{\spadesuit}$Johns Hopkins University \hspace{2.5mm} $^{\heartsuit}$University of Rochester \\
        \texttt{\{ksande25, rkriz1, detter2, vandurme\}@jhu.edu}}

\begin{document}

\maketitle

\begin{abstract}
\label{sec:abstract}
How are we able to learn about complex current events just from short snippets of video?
While natural language enables straightforward ways to represent under-specified, partially observable events, visual data does not facilitate analogous methods and, consequently, introduces unique challenges in event understanding. With the growing prevalence of vision-capable AI agents, these systems must be able to model events from collections of unstructured video data. To tackle robust event modeling in multimodal settings, we introduce a multimodal formulation for \textit{partially-defined} events and cast the extraction of these events as a three-stage span retrieval task. We propose a corresponding benchmark for this task, \dsetname, that consists of 14.5 hours of densely annotated current event videos and 1,168 text documents, containing 22.8K labeled event-centric entities. We propose a collection of LLM-driven approaches to the task of multimodal event analysis, and evaluate them on \dsetname. Results illustrate the challenges that abstract event understanding poses and demonstrates  promise in event-centric video-language systems.
\end{abstract}
\section{Introduction}
\label{sec:introduction}

\begin{figure}[t!]
\includegraphics[width=\linewidth]{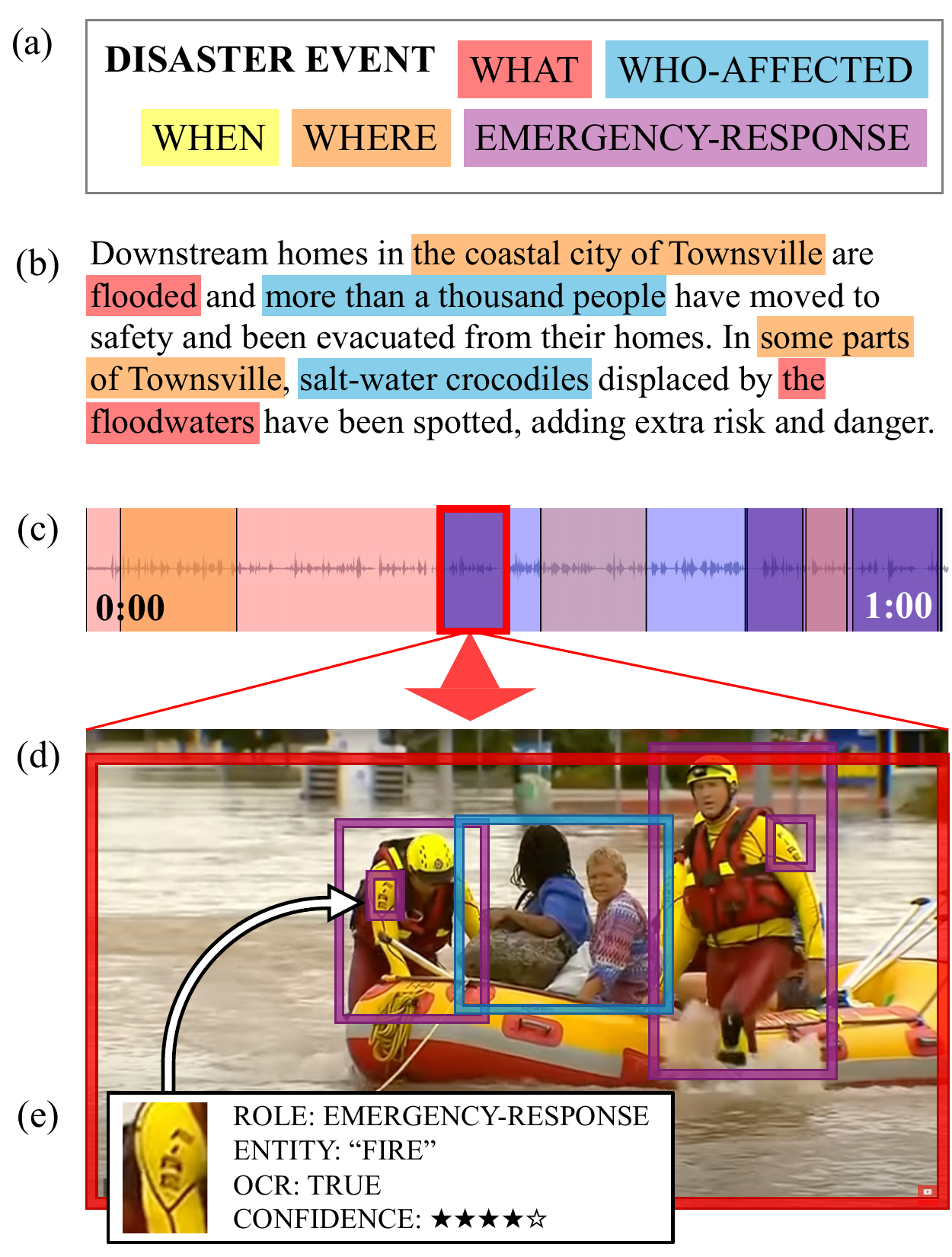}
\caption{In \dsetname\, every video-text pair is labeled with (a) an event template that guides the annotations, and annotations identifying entities that help fill these template roles at the (b) text, (c) temporal (video time stamps), and (d) spatial levels. Spatial annotations are also paired with (e) a natural language description of the visual content (or transcription of the readable text), a flag for whether the entity is natural language within the video, and confidence for how sure the annotator is that the entity relates to the event role in question.}
\label{fig:title}
\vspace{-1em}
\end{figure}

Event recognition is critical to how we understand the world. Evidence suggests that even pre-linguistic infants demonstrate a capacity for recognizing discrete events in real-world stimuli~\cite{wynn1996infants, yates2022neural}. Natural language reflects this cognitive inclination towards event modeling through the various linguistic structures tied to event representations~\cite{pustejovsky2005syntax, bohnemeyer2007principles}, and significant work considers how to model events conveyed through these structures formally~\cite{schein20138} and through data-driven approaches~\cite{li2022survey}.

In comparison, there is less work considering the relationship between these event models and the natural, temporal stimuli from which they are derived. A handful of tasks have been proposed, but rely on highly limited ontologies with events that can be fully depicted in seconds~\cite{chen2021joint, sadhu2021visual}. In contrast, event modeling systems for text can operate at a seemingly arbitrary level of event complexity, an ability enabled by the event-centric structure of natural language; for example, we can simply write a sentence like ``the War of 1812 lasted two years and eight months” while omitting the detail of what would theoretically be a two year and eight months-long video.

However, human understanding of most world events is founded in fractured snippets of this detailed, natural data. We regularly reason over intricate, recursive webs of multimodal information pieces, like visual-auditory data and language, to understand complicated situations. 
To work towards this same reasoning capability in AI systems, we consider the task of reasoning over noisy, multimodal data to extract information about events that are only partially shown in video content (temporally or spatially), or \textit{partially-defined} events. By shifting the focus towards grounding conclusions in specific pieces of data, we are also able to work towards reliable, transparent reasoning systems.

To this end, we formally describe the relationship between pieces of multimodal data and a pre-specified event, and propose a three-stage retrieval task for empirically modeling this relationship: These stages are modeled in \autoref{fig:title} sections (b), (c), and (d+e). We introduce an extension to the MultiVENT multilingual video retrieval dataset~\cite{sanders2024multivent}, MultiVENT-Grounded (\dsetname), to serve as a benchmark for this task. \dsetname{ } consists of over 14.5 hours of densely annotated multilingual video footage and 1,168 paired text documents, consisting of 22.8K multimodal event entity annotations. \autoref{fig:title} depicts a sample data point from \dsetname { }. We consider a collection of methods tackling each stage of this task, focusing on the employment of contemporary LLM and VLM models. We evaluate these approaches on our benchmark and provide an analysis of their comparative strengths and weaknesses.

In summary, we introduce:
\begin{enumerate}
\item \dsetname, the first open dataset for partially-defined event extraction in multimodal data, including multilingual content annotated in detail by professional linguists.
\item Experiments comparing approaches to partially-defined event extraction on \dsetname { }that demonstrate strong baseline results and illustrate the benefits and drawbacks of different contemporary modeling techniques on the task.
\end{enumerate}
\section{Related Work}
\label{sec:related-work}

\subsection{Events in Video}
\label{sub-sec:related-work:video-events}

\citet{tzelepis2016event} succinctly survey the scope of ``event-based media processing" in machine learning in the 2000's and early 2010's, covering how events are defined and represented in text, audio, and vision. They reference the standard understanding of events being changes in state~\cite{francois2005verl}, distinguishing it from definitions of more complex events like group actions~\cite{hakeem2004case}, news~\cite{sayyadi2009event}, or social settings~\cite{petkos2014social}. They go on to compare existing event models, differentiating them on the basis of relative and absolute temporal representation, relative and absolute spatial representation, and mereologic/causal/correlation inter-event relationship representation.

In the last decade, researchers have moved away from formal video event definitions, models, and extraction, instead focusing on downstream tasks that \textit{involve} event understanding, but do not model it as a task or evaluate for models' ability to perform it. Action recognition~\cite{herath2017going}, scene graph generation~\cite{DBLP:journals/corr/abs-2201-00443}, retrieval~\cite{spolaor2020systematic}, captioning~\cite{8698097}, Q\&A~\cite{zhong2022video}, and multipurpose benchmarks~\cite{wang2024internvid, huang2020movienet,lin2023videoxum,Grauman_2022_CVPR, patraucean2024perception} all involve events, but do not model them explicitly. In contrast to Tzelepis's organization of event models, \citet{sanders2024survey} argue that these implicit models generally diverge in terms of semantic and temporal complexity. 

\subsection{Event Extraction}
\label{sub-sec:related-work:event-extraction}

Event extraction can be broadly defined as the identification of specific information in unstructured data relating to changes of state. \citet{hogenboom2011overview} organize early event extraction work into data-driven approaches~\cite{lei2005system, liu2008extracting, okamoto2009discovering}, knowledge- (or expert-) driven approaches~\cite{nishihara2009event, aone2000rees, capet2008risk}, and hybrid approaches~\cite{lee2003ontology, jungermann2008enhanced}. Data-driven approaches resemble open-domain event extraction work, and knowledge-driven approaches resemble closed-domain work~\cite{liu2021overview}. They argue that data-driven methods do not sufficiently account for semantics, whereas knowledge-driven approaches are based on linguistic knowledge, but knowledge-driven methods introduce difficult modeling challenges stemming from the unstructured data involved. Meanwhile, hybrid approaches make up a large chunk of existing event extraction work: The structure of extracted events typically align with knowledge-driven methods, but the systems still take advantage of statistical approaches and the available data.

Multiple public event extraction challenges have been held over the years, including MUC~\cite{chinchor1997muc}, NIST's TDT~\cite{fiscus1999nist}, and ACE~\cite{doddington2004automatic}. These challenges involve extracting event ``templates" from unstructured text documents to produce structured knowledge bases, or closed-domain event extraction tasks. Closed-domain tasks tend to break a text down into a ``mention" span (describes the event), a ``trigger" span (single word that best describes the event), and ``argument" spans (entities that fill a specific event role). Following this setup, many closed-domain event extraction tasks like ACE are divided into four stages: Trigger detection, event identification, argument detection, and role identification. On the other hand, open-domain event extraction tasks are generally simpler, focusing on event detection and clustering. Text documents for both task types often focus on news articles and social media posts.

\subsection{Video Event Extraction}
\label{sub-sec:related-work:video-event-extraction}

While recent work in video understanding has generally avoided closed-ontology event extraction, there are a few notable exceptions: \citet{chen2021joint} introduce the task of MMEE, in which a system must map news clips to passages from their corresponding text articles through a closed-domain event template. They introduce the VM2E2 dataset to facilitate the study of this task. The events considered in this task are relatively simple, generally spanning a few seconds.

Meanwhile, \citet{sadhu2021visual} propose the task of VidSRL, which splits video clips into two-second segments and require models to (1) map the clips to event templates and (2) identify the relationships between the clips in terms of their event content. They introduce a corresponding dataset for this task consisting of movie clips. Again, these events typically span short time frames.

Our work differs from these efforts as it tackles partially-defined events depicted across multiple pieces of multimodal data, which as we explore in the paper, which introduces unique challenges that require new data and approaches to study effectively.
\begin{figure*}[t!]
    \centering
    \includegraphics[width=2\columnwidth]{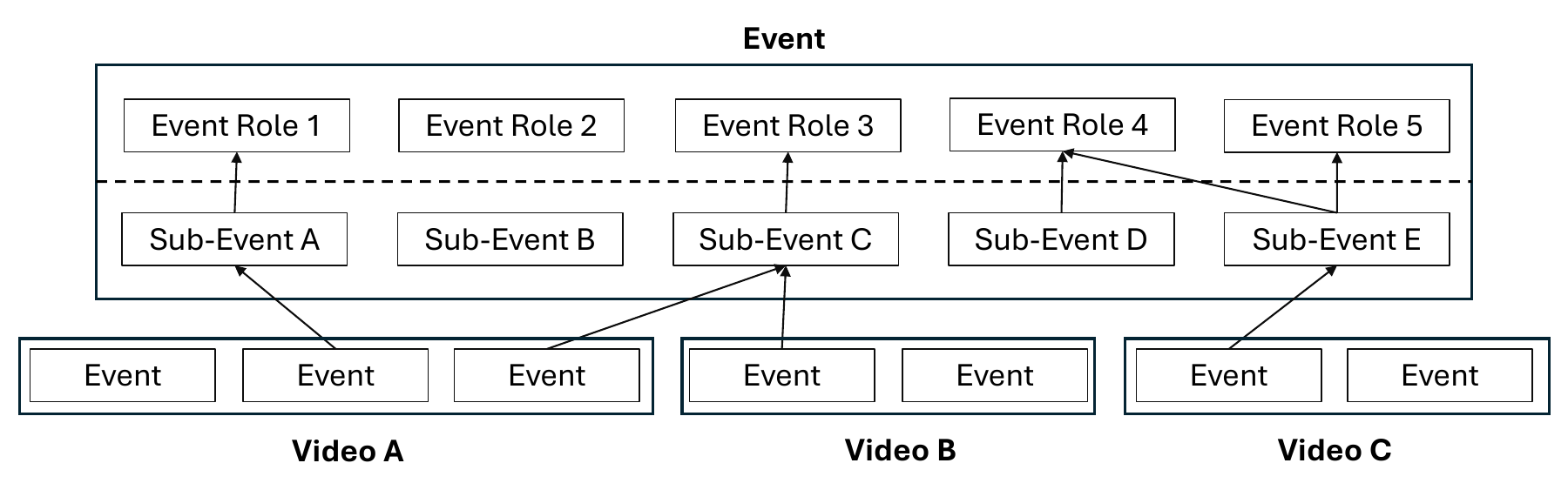}
    \caption{An illustration demonstrating the relationship between a pre-specified event and a collection of videos. Each role can simultaneously be defined by a set of event roles and by set of sub-events. Some subset of the sub-events characterize the role fillers, and some subset of the events depicted in a given video depict sub-events of the event.}
    \label{fig:model}
\end{figure*}

\begin{figure}[t!]
\includegraphics[width=\linewidth]{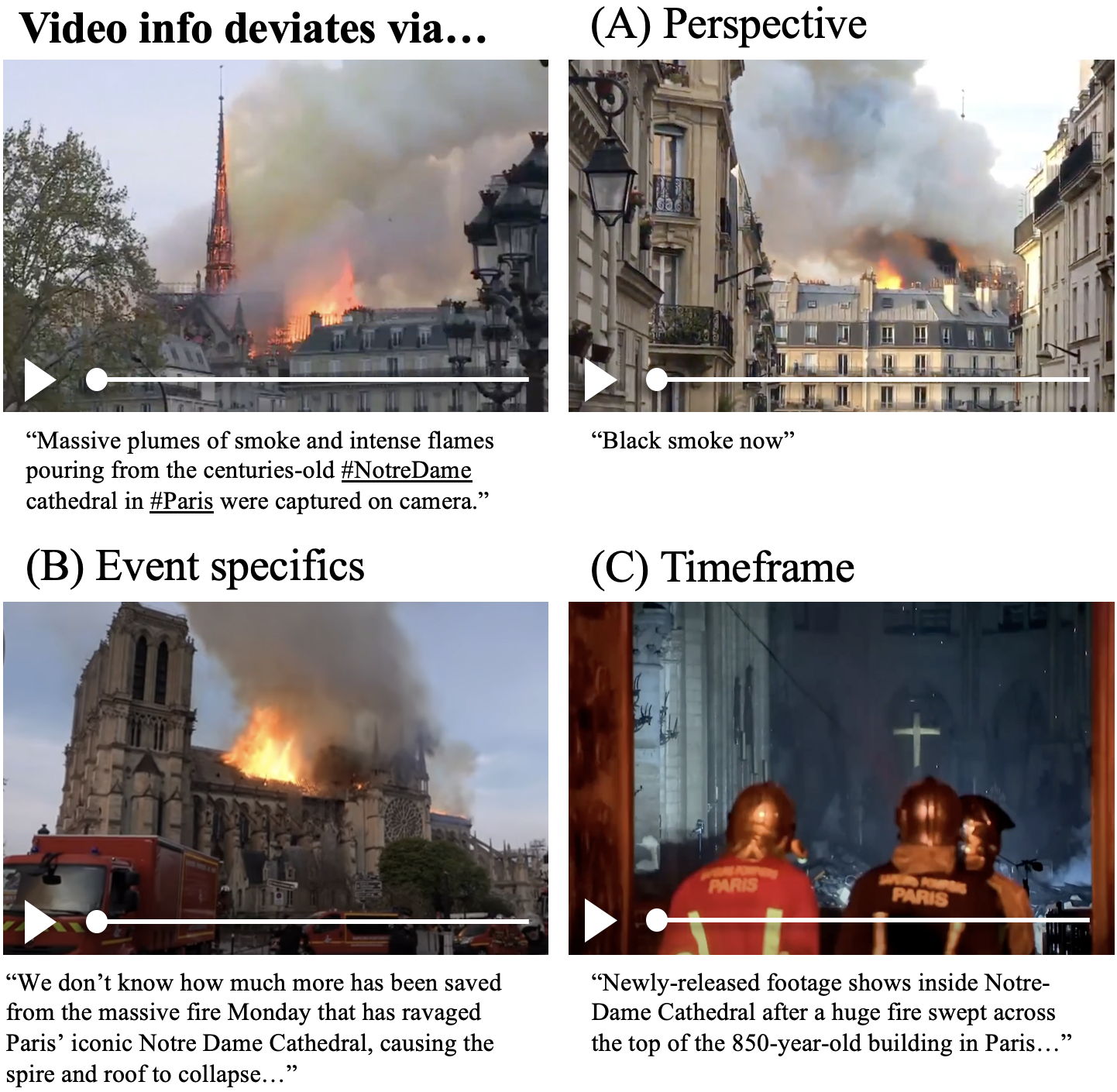}
\caption{Example video-description pairs depicting the Notre Dame Cathedral fire pulled from \dsetname. These videos illustrate how a single event can be described differently by different video clips: (A) provides the same semantic and temporal information as the reference video, but from a different perspective. (B) shows different semantic details, such as the fire trucks pictured at the bottom of the video frame. (C) was taken later than the reference video, showing a separate temporal snippet from the event in question with changed event semantics.}
\label{fig:partial}
\vspace{-1em}
\end{figure}

\section{Partially-Defined Event Extraction}
\label{sec:task}

\subsection{Partially-Defined Events}
\label{sub-sec:task:events}

Generally, data collection for video benchmarks falls into one of two categories: (1) A class of events is selected and individual instantiations of those events in video are found and labeled, or (2) videos are retrieved, and then events are found and annotated within them. The events annotated in both categories are wholly defined by the videos themselves, as they do not exist outside this piece of data. On the other hand, the events in MultiVENT exist outside the data in which they are depicted. During the dataset construction process, individual current events were identified, and multiple video-text pairs were retrieved that each depict different pieces of the event, temporally and spatially, and therefore contribute different information.

We consider the Davidsonian understanding of events~\cite{davidson1969individuation}; namely that (1) events are unique spatio-temporal entities and (2) they can be uniquely defined by a collection of attributes, known in event extraction literature as templates or frames. \citet{allen1984towards} goes further and describes an event as a non-homogeneous entity that exists at some place for some interval of time, meaning that events can be broken up into sub-parts. As individual data pieces in MultiVENT depict different pieces of the event, we can conclude that each provides a partial observation of some possibly empty subset of the sub-events that make up the event in question. We illustrate this concept in \autoref{fig:model}.

The idea of partial observations of sub-events is explored by Hwang et al. in their work on episodic logic~\cite{hwang1993episodic}. They argue that while an event can be \textit{defined} by one sentence, it may be \textit{described} by another. We adopt this formulation here: While videos in most benchmarks define events, the videos in MultiVENT describe and \textit{partially} define them. We explore the different ways videos can deviate in their partial descriptions through examples in \autoref{fig:partial}.

\subsection{Task Formulation}
\label{sub-sec:task:task-formulation}

So, it doesn't make sense to formulate event understanding over collections of data as a traditional event extraction problem. Instead of modeling an event as it is defined by data, we want to model it as it is described by data. We cast this as a retrieval problem, where we aim to answer the question of how data influences our understanding of an event defined through the inputted template.

\paragraph{Input}
Specifically, our input is a collection of some sample of text and video data of nonzero size, and an event template which guides the event modeling. We treat the event template as the retrieval query - we query on a specific event type, defined by the template.

\paragraph{Output}
We aim to return the set of ``descriptive spans" present in our input data that describe the event dictated by the inputted template. These spans are both textual and spatio-temporal, and so we explicitly decompose the event understanding task into three distinct stages:

\begin{enumerate}
\item Text span retrieval: Retrieve the set of descriptive spans within the text samples that describe the event under consideration, labeled with the event role they describe.
\item Temporal span retrieval: Retrieve the set of video intervals, described as start-end points, that describe the event under consideration, labeled by role.
\item Spatial span retrieval: Given an additional input variable of time point within a video, retrieve the set of spatial bounding boxes that mark the visual entities that describe the event. These should also be labeled with the template role they correspond to.
\end{enumerate}

As shown above, we do not explicitly include template filling as part of the task. This is because, as the events are not defined by the input data, there is no meaningful function to map between input content and the true underlying labels that does not involve modeling for substantial unknowable noise. We instead focus on retrieving useful information within the provided data that can then be used by systems to make meaningful hypotheses about the underlying event, better matching how humans learn and understand partially-defined events.

\subsection{Metrics}
\label{sub-sec:task:metrics}

For this task, we consider five metrics: Precision, recall, F1, CEAF-RME, and IoU. We outline these and their implementations below.

\paragraph{Precision, recall, and F1}
We first implement span-based generic retrieval scores, in which we consider the granular units of each domain (characters, seconds, and pixels) as the entities over which we are doing retrieval. Let $R_f$ be the set of ground truth entities for role filler
$f$, and $S_f$ be the set of predicted entities. Then, our corresponding metrics can be formalized as

$$P_0=\frac{1}{|F|}\sum_{f\in F}\frac{|R_f\cap S_f|}{|S_f|}\hspace{4mm}R_0=\frac{1}{|F|}\sum_{f\in F}\frac{|R_f\cap S_f|}{|R_f|}\hspace{4mm}$$

$$F_0=\frac{2P_0R_0}{P_0+R_0}$$

\paragraph{CEAF-RME} For text evaluation, we implement \citet{chen2022iterative}'s modification to \citet{du2020grit}'s template filling metric CEAF-REE, CEAF-RME. CEAF-RME uses the Kuhn-Munkres maximum bipartite matching algorithm~\cite{kuhn1955hungarian, munkres1957algorithms} to map predicted spans to ground truth annotations and then report the resulting precision, recall, and F1 scores. However, it relaxes the definition of ``map" to allow for partially overlapping entities.








\paragraph{Role-Filling IoU}
For temporal and spatial evaluations, we implement 2D and 3D versions of IoU at 0.5, 0.7, and 1, treating the temporal/spatial annotations for one role as a single entity annotation. We define our metric as

$$R\,(\text{IoU}=k)=\frac{1}{|F|}\sum_{f\in F}\min \left(1,\,\frac{|R_f\cap S_f|}{k\,|R_f\cup S_f|}\right)$$
\section{\dsetname}
\label{sec:data}

\begin{figure*}[t!]
    \centering
    \includegraphics[width=2\columnwidth]{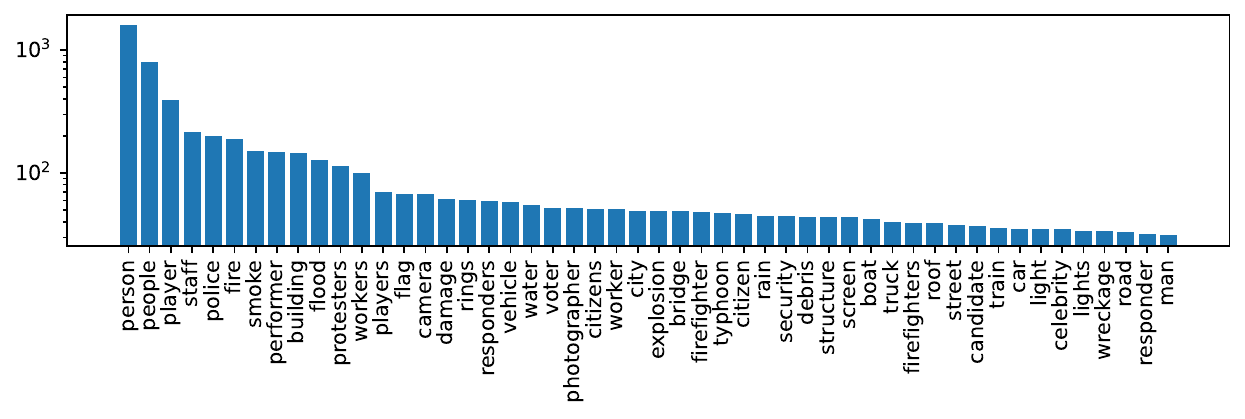}
    \caption{The distribution, using a logarithmic scale, of spatial entity labels (post-data cleaning) in \dsetname. The labels reflect the domain of annotated content, e.g., as many videos depict emergencies, ``police" and ``fire" labels are common.}
    \label{fig:common_nouns}
    \vspace{-1em}
\end{figure*}

\subsection{Videos}
\label{sub-sec:data:videos}

We select a 1,168 video subset of the MultiVENT dataset, a collection of multi-format, multilingual videos depicting 260 current events paired with natural language descriptions and aligned Wikipedia articles. The mean length of these videos is 83.7 seconds. Our selected videos are distributed across five languages (Arabic, Chinese, English, Korean, and Russian) and four news genres (emergencies, political content, social events, and science/business coverage). The videos are organically produced for speakers of these languages, and can include multiple videos (up to ten) depicting the same event.

\subsection{Event Information}
\label{sub-sec:data:event-info}

We first align each depicted event to the FrameNet template~\cite{baker1998berkeley} that provides the best semantic match. We consolidate these templates into a set of seven categories that sufficiently covers the span of event types: Emergencies, elections, phenomenon launch or discovery, political development, political demonstrations, social events, and sports events. We adapt these templates to the visual domain by removing event roles that cannot be immediately shown in visual content, and occasionally adding roles that are important to understanding the events visually. For each text-video pair in our MultiVENT subset, we ground the corresponding event roles in three dimensions: Text, time, and space. Below, we describe this grounding in detail, and provide the full ontology in \autoref{tab:multivent-ontology}. 

\paragraph{Textual annotations.} 
We map each event role to the (possibly empty) subset of continuous spans within the text description string that provide information about the event role in question.

\paragraph{Temporal annotations.}
We identify time spans within the video content during which a visible entity, entities, or interaction between entities provides information about the event role. We identify the start and end times, and if the same entity reappears later in the video, we annotate that time point as a separate segment mapping to the same role. We map these time segments to their corresponding event role labels.

\begin{table}
    \centering
    \small
    \caption{Examples of spatial entity labels before and after data cleaning using GPT, done with the goal of making long human-written labels concise and uniform with respect to the rest of the dataset.}
    \begin{tabular}{p{4.5cm}l}
    \toprule
        \textbf{Original label} & \textbf{New label} \\
         \hline
        policeman, security, or another federal official dressed in similar attire & law enforcement\\
        \hline
        lights flickering, presumable due to shaking & flickering lights\\
        \hline
        residents of the area that was attacked going through the rubble & residents in rubble\\
        \hline
        grandson helping grandmother to safety & family evacuation\\
        \hline
        a firetruck plowing through a flood & firetruck in flood\\
        \bottomrule
    \end{tabular}
    \label{tab:shortened}
    \vspace{-1em}
\end{table}

\begin{table}
\small
    \centering
    \small
    \caption{Distribution of annotated video types in \dsetname, partitioned by language and semantic domain. As shown, the most videos are English, but this set still makes up a minority of the dataset in total.}
    \begin{tabular}{ccccccc}
    \toprule
        \textbf{Template} & \textbf{AR} & \textbf{ZH} & \textbf{EN} & \textbf{KO} & \textbf{RU} & \textbf{Total} \\
        \midrule
        Emergency & 99  &  69 &  121 & 118  & 81  & 487 \\
        \midrule
        Election & 0  & 17  & 23  & 20  & 5  & 65 \\
        Political & 17  & 34  & 60  & 33  & 33  & 177 \\
        Demonstration & 9  & 30  &  31 &  18 & 29  & 117 \\
        \midrule
        Social &  0 & 33  &  58 & 20  & 21  & 132 \\
        Sports & 0  & 6  & 50  & 0  & 5  & 61 \\
        \midrule
        Discovery & 0  & 45  & 71  & 0  & 13  & 129 \\
        \midrule
        \textbf{Total} & \textbf{125} & \textbf{234} & \textbf{414} & \textbf{208} & \textbf{187} & \textbf{1168} \\
        \bottomrule
    \end{tabular}
    \label{tab:video-counts}
    \vspace{-1em}
\end{table}

\begin{table}
\small
    \centering
    \caption{Statistics corresponding to the raw annotations before and after post-processing (but after quality filtering).}
    \begin{tabular}{ccccc}
    \toprule
        \textbf{Annotation} & \textbf{25\%} & \textbf{50\%} & \textbf{75\%} & \textbf{Max} \\
         \midrule
         \multicolumn{5}{c}{{Annotation Counts per Video}}\\
         \midrule
        Text & 2 & 5 & 9 & 105\\
        Temporal & 2 & 4 & 10 & 223\\
        Spatial (Vis) & 1 & 2 & 7 & 249\\
        Spatial (OCR) & 0 & 2 & 7 & 72\\
        \textbf{Text + Spatial} & \textbf{7} & \textbf{14} & \textbf{23} & \textbf{251}\\
        \midrule
        \multicolumn{5}{c}{{Span Sizes}}\\
        \midrule
        Text (char) & 5 & 9 & 17 & 180\\
        Temporal (sec) & 2 & 4 & 8 & 186\\
        Spatial width (\%) & 11 & 27 & 59 & 100\\
        Spatial height (\%) & 22 & 42 & 70 & 100\\
        OCR (char) & 4 & 7 & 13 & 182\\
        \midrule
        \multicolumn{5}{c}{{Post-Processed Statistics}}\\
        \midrule
        Temporal count & 1  & 3  & 6  & 32 \\
        Temporal length (sec) & 4  & 7  & 19  & 186 \\
        \bottomrule
    \end{tabular}
    \label{tab:annotation-stats}
\end{table}

\paragraph{Spatial annotations.}
To spatially ground the event-relevant visual entities, for every temporal span-event role pair we select one representative frame from the temporal span in which the visual entity or entities are clearly visible, and draw a bounding box around the relevant content. If two entities corresponding to the same role contain negligible minimal space between them, they may occupy the same bounding box.

\paragraph{Entity metadata.}
For each spatial entity annotated in the previous section, alongside the event role name we assign a set of corresponding metadata labels to provide additional semantic information. Specifically, we include (1) a short, natural language description of the entity. This is recorded in English unless the entity is OCR content, in which case it is written as a direct transcription in the language it appears in. (2) We then note whether or not the entity in question is OCR content. (3) We include a human confidence score indicating how certain a human is that the entity in question is related to the corresponding event role. We include this confidence score because visual content is often ambiguous, and work suggests that human confidence scores are often a sound method for quantifying the clarity of such data~\cite{sanders2022ambiguous}. This way, we avoid producing overly conservative annotations that miss key information that could help a model to identify an event.

\subsection{Annotations}
\label{sub-sec:data:annotations}

We recruit a team of professional linguists to annotate the video content. First, we train annotators through an hour-long seminar describing the task, and then another hour-long tutorial session during which annotators annotate a test video-text pair with ``gold" annotations determined by the authors. This test annotation task is returned to the authors to be scored, and feedback is provided. Once annotators annotate the test task with a sufficient overall F1 score compared to the gold annotations, they are granted access to begin annotating videos in their preferred language. Annotators are encouraged to ask each other and the authors questions, and they are provided with a 20-page annotation manual and tutorial videos as reference material. Main instructions from the manual describing the task are included in \autoref{appendix:instructions}. All videos are annotated up to the 60-second mark.

\begin{figure*}[t!]
    \centering
    \includegraphics[width=2\columnwidth]{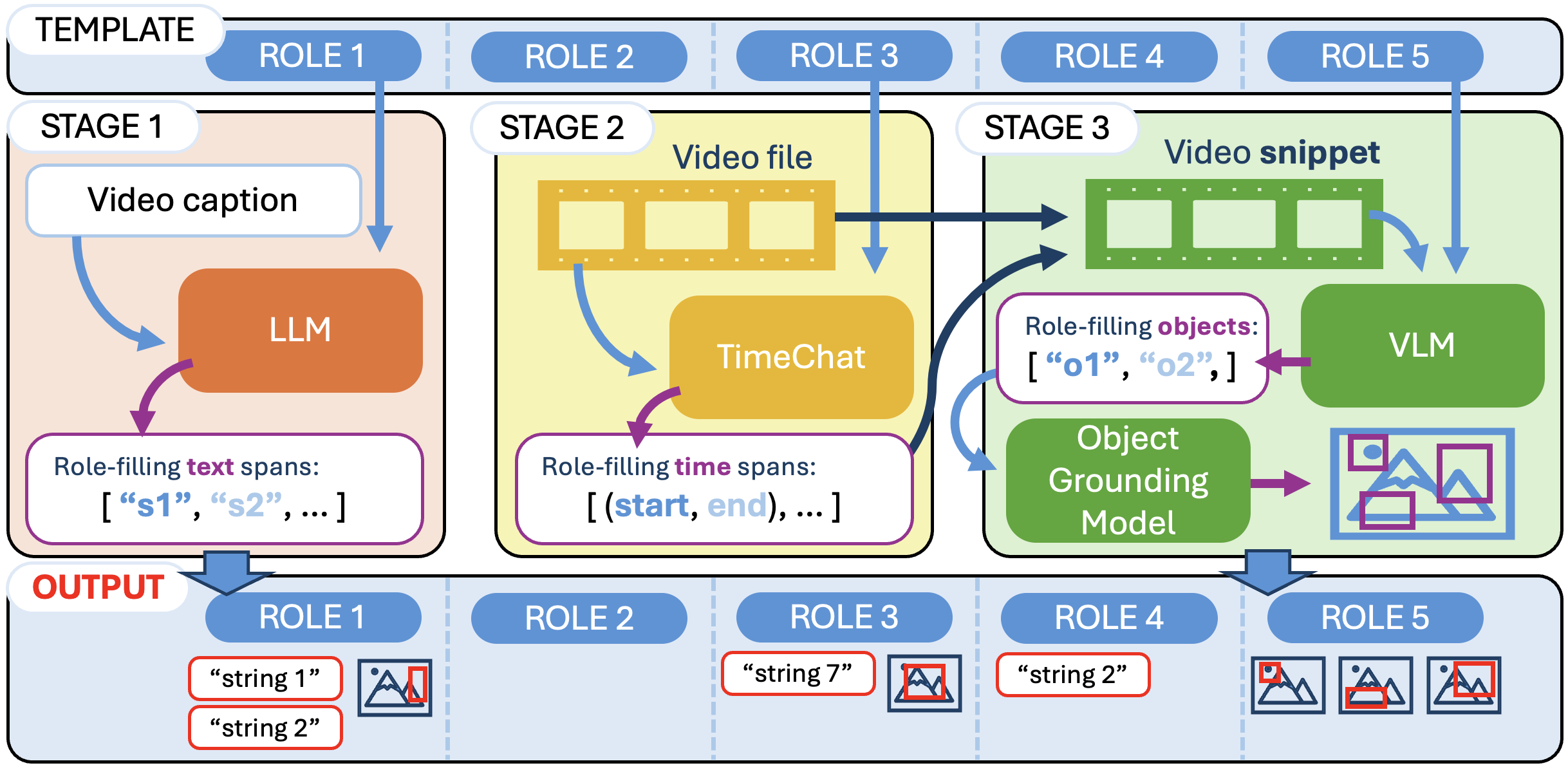}
    \caption{The complete partially-defined event understanding task, broken down into three stages. Stage 1 retrieves relevant text spans, stage 2 retrieves relevant temporal spans, and stage 3 retrieves relevant spatial spans. The output of computing these sub-tasks in sequence is a filled event template using both textual and spatio-temporal evidence from the video-language data.}
    \label{fig:task}
\end{figure*}

\subsection{Data Cleaning}
\label{sub-sec:data:data-cleaning}

The natural language descriptions labeling spatial spans have different levels of detail depending on the annotator. After stripping whitespace and converting characters to lowercase, there are 1964 distinct labels for 8299 visual entities. The 90th percentile length is 23 characters, but the maximum is 168 characters. The 90th percentile word count is 2, but the maximum is 29. To adjust these longer descriptions to better serve as entity labels for classification, we use GPT-3.5 to reduce the length of each description over 3 words, accounting for 487 distinct labels. We take the first phrase returned by GPT. We include a sample of the replaced descriptions in \autoref{tab:shortened}. The resulting set consists of 2484 distinct labels. The distribution of spatial labels is illustrated in \autoref{appendix:prompts}, and we include the prompt used for cleaning in \autoref{prompt:text_cleaning}.

We also aggregate timeline annotations such that, for any label, two overlapping or adjacent timeline annotations are merged into a single segment.

\subsection{Dataset Statistics}
\label{sub-sec:data:data-stats}

In total, our dataset consists of 8.4K textual annotations, 4.9K temporal annotations, and 14.4K spatial annotations (8.3K purely visual and 6.1K OCR) for a total of 27.6K labels, or 22.8K labeled event-relevant entity instances. We include details regarding annotated video categories and annotation statistics in \autoref{tab:video-counts} and \autoref{tab:annotation-stats}. 
\section{Experiments}
\label{sec:experiments}

The three-stage task introduced in Section \ref{sec:task} is depicted in \autoref{fig:task}. The task is difficult, as it requires extracting complex information from multiple modalities across highly varied domains. Therefore, we propose a collection of approaches that leverage the generalizability and high-level reasoning abilities of LLMs and VLMs to the three stages of the task. Below we provide an overview of their implementation and differences, as well as quantitative results comparing them on the three task stages using \dsetname.


\begin{table}
    \small
    \centering
    \caption{Performance of various models evaluated on the text span retrieval task. We report precision, recall, and F1 scores and the CEAF-RME metrics as described in \autoref{sec:task}. \textdaggerdbl{} indicates the model was prompted with an in-context learning prompt created independently of the \dsetname{} dataset.}
    \begin{tabular}{ccccccc}

    \toprule
        \textbf{Method} & \textbf{P$_0$} & \textbf{R$_0$} & \textbf{F$_0$} & \textbf{P$_C$} & \textbf{R$_C$} & \textbf{F$_C$}\\
        \midrule
        LLaMA 2 & 79.1 & 13.8 & 23.4 & 56.3 & 13.3 & 21.6 \\
        GPT-3.5 & 83.3 & 43.7 & 57.2 & 74.7 & 38.2 & 50.6 \\
        GPT-4o & 83.3 & \textbf{56.4} & \textbf{67.2} & 77.1 & \textbf{51.5} & \textbf{61.8} \\
        LLaMA 2$^\text{\textdaggerdbl}$ & 66.7 & 21.3 & 32.3  & 43.3 & 20.4 & 27.8 \\
        GPT-3.5$^\text{\textdaggerdbl}$ & 77.8 & 46.4 & 58.1 & 72.0 & 42.6 & 53.5 \\
        GPT-4o$^\text{\textdaggerdbl}$ & \textbf{84.1} & 51.6 & 63.9  & \textbf{79.1} & 48.1 & 59.8 \\
        \bottomrule
    \end{tabular}
    \label{tab:text-exps}
\end{table}

\begin{table}[t!]
   \small
    \centering
    \caption{Performance of various models evaluated on the temporal span retrieval task.}
    \begin{tabular}{cccccc}
    \toprule
        & \textbf{P$_0$} & \textbf{R$_0$} & \textbf{F$_0$} & \textbf{IoU$_{.5}$} & \textbf{IoU$_{.7}$}\\
        \midrule
        TC & 45.67 & 28.77 & 27.53 & 37.07 & 29.73\\
        TC-C & \textbf{50.51} & \textbf{32.38} & \textbf{33.36} & \textbf{47.37} & \textbf{37.44}\\
        TC-VG & 47.50 & 24.72 & 25.66 & 35.00 & 27.25 \\
        \bottomrule
    \end{tabular}
    \label{tab:temporal}
\end{table}

\begin{table}[t!]
    \small
    \centering
    \caption{Performance of the detected entity mapping and caption grounding approaches on the spatial span retrieval task. We report Role Filling IoU as described in \autoref{sec:task}, and additionally the modified CEAF-RME metrics described in \autoref{sub-sec:exps:spatial}, P$_{C*}$/R$_{C*}$/F$_{C*}$. IV denotes InternVL, L denotes LLaVA, D denotes Grounding-DINO, and G denotes GLIP. P, R, and F scores do not change with grounding models as they only measure the quality of the captioners.}
    \begin{tabular}{ccccccc}
    \toprule
        \textbf{Method} & \textbf{P$_{C*}$} & \textbf{R$_{C*}$} & \textbf{F$_{C*}$} & \textbf{IoU$^\prime_{.3}$} & \textbf{IoU$^\prime_{.5}$} & \textbf{IoU$^\prime_{.7}$} \\
        \midrule
        IV2B+D & \textbf{78.1} & 14.9 & 25.0 & 3.5 & 3.0 & 2.6 \\
        IV2B+G & \textbf{78.1} & 14.9 & 25.0 & 8.5 & 7.3 & 6.4 \\
        IV4B+D & 58.8 & 30.3 & \textbf{40.0} & 7.9 & 6.7 & 5.8 \\
        IV4B+G & 58.8 & 30.3 & \textbf{40.0} & 15.2 & 13.1 & 11.4 \\
        L7B+D & 52.9 & \textbf{31.4} & 39.5 & 15.2 & 13.4 & 12.0 \\
        L7B+G & 52.9 & \textbf{31.4} & 39.5 & \textbf{27.2} & \textbf{23.8} & \textbf{21.1} \\
        \bottomrule
    \end{tabular}
    \label{tab:video-based}
    \vspace{-1em}
\end{table}


\begin{table}[t!]
    \small
    \centering
    \caption{OCR results. Reported numbers are the retrieval rates for 50\%+ and 100\% of the relevant strings annotated in the videos, partitioned by writing system.}
    \begin{tabular}{cccccc}
    \toprule
        & \textbf{AR} & \textbf{EN} & \textbf{KO} & \textbf{RU} & \textbf{ZH} \\
        \midrule
        \multicolumn{6}{c}{\textbf{IoU}$_{0.5}^*$}\\
    \midrule
        Paddle & 4.59 & 78.94 & 47.68 & 25.66 & 84.81 \\
        EasyOCR & 74.23 & 75.27 & 76.20 & 69.45 & 73.51 \\
        \midrule
            \multicolumn{6}{c}{\textbf{IoU}$_{1.0}^*$}\\
    \midrule
            Paddle & 0. & 60.17 & 36.09 & 08.33 & 55.69 \\
        EasyOCR & 51.03 & 55.44 & 52.96 & 41.10 & 47.85 \\
        \bottomrule
    \end{tabular}
    \label{tab:ocr}
    \vspace{-1em}
\end{table}

\subsection{Text Span Retrieval}
\label{sub-sec:exps:text}
We cast the span extraction problem as an instance of the original dataset annotation task. Provided with the text description and the template, we ask the model to retrieve each segment of text that helps to answer the provided template role. \footnote{We include all LLM prompts in the appendix.} We attempt a zero-shot and in-context learning versions of the task. For the in-context learning task, for each queried event template role we provide two sample outputs taken from data not included in \dsetname. For this approach, we test GPT-3.5, GPT-4o~\cite{achiam2023gpt}, and LLaMA 2 7B~\cite{touvron2023llama}.

We evaluate these methods on the task of text evidence retrieval, and report results in \autoref{tab:text-exps}. We find that most of the LLMs perform relatively well on the task, and do not benefit substantially from few-shot prompting. For some of the models, few-shot prompting hurts performance. High recall possibly indicates longer retrieved spans, potentially full sentences, whereas precision indicates that single-word spans were likely more frequent. 

\subsection{Temporal Span Retrieval} 
\label{sub-sec:exps:temporal}
Temporal span retrieval is difficult in that it requires the model to return time stamps, which is challenging for most systems. We introduce a suite of finetuned video language models based on the TimeChat \cite{ren2024timechat} architecture. We train TimeChat on a variety of settings: TimeChat Base (TC) -- the original model from \cite{ren2024timechat}, TimeChat Charades (TC$_c$) -- the original checkpoint tuned on Charades \cite{zhang2019temporal}, and TimeChat Video Grounding (TC$_{vg}$), tuned on four temporal video grounding datasets \cite{hendricks2017localizing,zhang2019temporal,oncescu2021queryd, zala2023hierarchical}.

We evaluate these methods on the task of temporal retrieval using the \dsetname{} dataset. We report results in \autoref{tab:temporal}. We find that VideoLLMs are capable of generalizing to the temporal retrieval task when it matches the task of temporal video grounding observed during instruction tuning. We see the best model in performance when the VideoLLM undergoes additional instruction tuning for the Charades \cite{zhang2019temporal}. However, when the VideoLLM is tuned further on additional temporal video grounding data, it loses its ability to generalize to \dsetname. 
Traditional temporal video grounding focuses on entity-specific tasks, but the \dsetname{} ontology covers a broader event spectrum, some of which are not human-centric and out-of-distribution from the training data. We offer a further breakdown of the performance of the temporal retrieval task across the languages and template types in \autoref{sub-appendix:additional-results:temporal}. 

\subsection{Spatial Span Retrieval} 
\label{sub-sec:exps:spatial}
We employ a two-step process for spatial span retrieval. (1) Given a frame from the video and an event template, we instruct a VLM to identify short phrases within the image that help to answer the template questions. We test LLaVA 1.5 7B~\cite{liu2023improved} and two sizes of InternVL~\cite{chen2024internvl} for this step. (2) Then, given the descriptive phrases output by the VLM, we use a phrase grounding model to bound the relevant entities in the image. For this, we use MM-Grounding-DINO~\cite{zhao2024open} and a GLIP architecture~\cite{Li_2022_CVPR} trained on a collection of detection datasets~\cite{shao2019objects365, kamath2021mdetr, sharma2018conceptual, ordonez2011im2text}.

Through the annotation protocol, one frame is annotated per entity, but not all entities are annotated in the same frame. Therefore, not all relevant information is necessarily labeled in any individual frame. To account for this, for this evaluation we modify our role-filling IoU metric such that the IoU is computed per mapped bounding box, mapping from the top-3 bounding box predictions, and denote this as IoU$^\prime$. In addition to the role-filling metric, for the captioning models we additionally compute a similarity-based retrieval metric, similar to CEAF-RME, that for each role in a video computes how similar the retrieved entity captions are to the ground truth natural language entity captions across all frames. For this, we compute the CEAF-RME for the retrieved labels, but replace the span similarity metric $\phi$ with the output of a distilled NLU model trained on sentence pairs that computes the semantic similarity between labels.

Results are shown in \autoref{tab:video-based}. Compared to the precision and recall scores of the more traditional textual information extraction task, the VLM captioners lag slightly behind in performance. The performance drop is noticable when shifting to object grounding---this is unsurprising as the models are both smaller than VLMs and generally trained on a more limited collection of datasets. 

\subsection{OCR Analysis}
As an additional experiment, in an attempt to identify how well relevant OCR information in the videos can be extracted by contemporary systems we evaluate PaddleOCR~\cite{du2020pp} and EasyOCR~\cite{shi2015end, baek2019character} on the annotated frames of \dsetname. We compute the recall for all retrieved strings that (1) overlap in bounding boxes with the ground truth and (2) correctly retrieve either at least 50\% (IoU$_{0.5}^*$) or all (IoU$_{1.0}^*$) of the ground truth string. As OCR for multiple languages can appear in the same video, we only compare the predictions against strings written with the same writing system.

Results are reported in \autoref{tab:ocr}. EasyOCR is able to retrieve approximately 50\% of the full relevant strings per language, indicating that current OCR systems are capable of retrieving a significant amount of event-centric text, and that there is still room for improvement.
\section{Conclusion}
\label{sec:conclusion}

In this paper we explore the problem of partially-defined event extraction in noisy, multimodal content, and propose a multimedia event formulation for tackling it. We subsequently introduce \dsetname, a corresponding three-stage event extraction benchmark consisting of densely annotated news content building on the MultiVENT dataset, and a set of potential methods for peforming the task stages. We report method performance on this benchmark, comparing zero-shot/few-shot LLM based methods to more traditional fine-tuned approaches. This paper marks an important step towards robust AI agents that can synthesize complex amalgamations of natural stimuli and processed information with human-level ability.

In future work, we hope to develop comprehensive systems that can address each stage of the event extraction task in tandem, allowing the outputs for each stage to be conditioned on the intermediate states and outputs of the other two stages. We believe that incorporating multilingual OCR systems into our pipeline would also improve performance, especially for videos not taken in predominantly English-speaking countries. Finally, while we introduce a preliminary exploration of partially-defined events in visual data, there is a wide range of ideas left to explore further on this topic, with the goals of better performance on event-centric vision benchmarks and a better understanding of human event understanding.
\section*{Limitations}
Below, we consider limitations and ethical considerations within the paper.
\label{sec:limitations}
\paragraph{Dataset}
The \dsetname{} dataset has been carefully annotated by speakers of the five target languages, but annotators are not immune to human error. Ideally, we would include multiple rounds of quality checks to ensure that all annotations are accurate. Furthermore, some aspects of the annotation task are subjective in nature, as complex visual data is inherently noisy. We attempt to account for this by including confidence judgments for spatial annotations. The \dsetname{} dataset is smaller than some contemporary video datasets, due to the intensive time and monetary costs of annotating a dataset at a sufficient level of detail. We argue that as foundational models continue to grow in prominence, small, well-annotated evaluation datasets will become critical for analyzing the performance of such systems on complex tasks.

\paragraph{Experiments}
VideoLLMs, like TimeChat \cite{ren2024timechat}, were selected as the baseline for temporal span retrieval due to their strong ability to generalize to unseen data. However, it is common consensus in the literature that LLMs struggle with the tokenization of numeric data \cite{golkar2023xval, singh2024tokenization, spathis2023step}. \citet{ren2024timechat} also report the difference in performance between TimeChat and task-specific and video-centric models between 5-7 pts in R@1 IoU. We leave the exploration of stronger video grounding of VideoLLMs and task-specific models to future work.

The experiments of the three task stages were conducted independently due to the complexity and general difficulty of the task. In future work, we hope to develop end-to-end systems that can simultaneously address these three task stages.

\noindent As with any task involving real-world content, there is potential for abuse. We strongly encourage researchers to be mindful of AI system biases that may arise when evaluating on or using this dataset to build systems.\\

\noindent The original MultiVENT dataset is licensed under CC-BY 4.0. Our work builds on the ideas introduced in the MultiVENT publication, in the same spirit of multilingual, multimodal information retrieval and extraction.

\section*{Acknowledgments}
\small
We thank Frank Ferraro for his contributions to the task formulation. Alexander Martin was also supported in part by the Goergen Institute for Data Science at the University of Rochester. We would also like to thank the many annotators who contributed to the dataset.


\bibliography{custom}

\appendix
\section{Additional Results}
\label{appendix:additional-results}




\subsection{Supplementary Temporal Evaluations}
\label{sub-appendix:additional-results:temporal}
In this section we discuss the results of the experiments across the multiple languages of \dsetname. 

\paragraph{Language} 
In \autoref{tab:temporal-language}, there is not a large variety in deviation in scores across the language types (with the exception of Korean performing 3-5 points worse than other languages). These results are consistent across each trained model. We interpret this result to signal that the visual content of the videos must be similar across languages. 

\paragraph{Event Type}
In \autoref{tab:temporal-type}, we do not observe the same consistency between event types that we do with languages. The best result across each model is the social split including social and sporting events. This result is intuitive as sporting and social events are human-centric activities often occurring in the training data. Meanwhile, science and technology news videos tend to involve highly abstract visual content paired with OCR information. More investigation into incorporating OCR into temporal modeling methods may provide further insight. 
\begin{table}
    \centering
    \begin{tabular}{cccccc}
        \toprule
         &  & TC & TC$_c$ & TC$_{vg}$ \\
         \midrule
        \multirow{3}{*}{$\textbf{EN}$} 
        & F1 & 30.32 & 33.73 & 26.57\\
        & IoU$_{0.5}$ & 41.67 & 48.51 & 37.15\\
        & IoU$_{0.7}$ & 34.04 & 38.93 & 29.09\\
        \midrule
        \multirow{3}{*}{$\textbf{RU}$} 
        & F1 & 28.83 & 33.31 & 26.76\\
        & IoU$_{0.5}$ & 37.34 & 46.41 & 35.97\\
        & IoU$_{0.7}$ & 29.98 & 35.95 & 27.79\\
        \midrule
        \multirow{3}{*}{$\textbf{CH}$} 
        & F1 & 25.74 & 34.03 & 25.07\\
        & IoU$_{0.5}$ & 35.59 & 48.41 & 34.34\\
        & IoU$_{0.7}$ & 27.80 & 38.03 & 26.47\\
        \midrule
        \multirow{3}{*}{$\textbf{KO}$} 
        & F1 & 23.07 & 30.83 & 23.11\\
        & IoU$_{0.5}$ & 30.08 & 42.70 & 30.31\\
        & IoU$_{0.7}$ & 23.63 & 33.07 & 23.76\\
        \midrule
        \multirow{3}{*}{$\textbf{AR}$} 
        & F1 & 36.65 & 35.38 & 26.34\\
        & IoU$_{0.5}$ & 34.69 & 51.04 & 35.17\\
        & IoU$_{0.7}$ & 28.04 & 41.13 & 27.44\\
        \bottomrule
    \end{tabular}
    \caption{Temporal Span Results By Language}
    \label{tab:temporal-language}
\end{table}

\begin{table}
    \centering
    \begin{tabular}{cccccc}
        \toprule
         &  & TC & TC$_c$ & TC$_{vg}$ \\
         \midrule
        \multirow{3}{*}{$\textbf{P}$} 
        & F1 & 23.85 & 34.76 & 23.27\\
        & IoU$_{0.5}$ & 49.43 & 48.51 & 32.04\\
        & IoU$_{0.7}$ & 38.69 & 38.93 & 24.55\\
        \midrule
        \multirow{3}{*}{$\textbf{S}$} 
        & F1 & 33.39 & 36.69 & 30.07\\
        & IoU$_{0.5}$ & 52.09 & 46.41 & 41.11\\
        & IoU$_{0.7}$ & 42.03 & 35.95 & 32.84\\
        \midrule
        \multirow{3}{*}{$\textbf{E}$} 
        & F1 & 29.17 & 32.61 & 26.47\\
        & IoU$_{0.5}$ & 46.14 & 48.41 & 35.60\\
        & IoU$_{0.7}$ & 36.34 & 38.03 & 27.59\\
        \midrule
        \multirow{3}{*}{$\textbf{T}$} 
        & F1 & 22.98 & 26.79 & 22.63\\
        & IoU$_{0.5}$ & 32.31 & 27.94 & 31.71\\
        & IoU$_{0.7}$ & 26.00 & 24.76 & 24.84\\
        \bottomrule
    \end{tabular}
    \caption{Temporal Span Results By Event Type. P - Political, S - Social, E- Emergency, T - Technology}
    \label{tab:temporal-type}
\end{table}


\section{LLM and VLM prompts}
\label{appendix:prompts}
In this section, we provide the collection of prompts used for data cleaning and experiments outlined in \autoref{sec:experiments}. The prompts listed are:
\begin{enumerate}
\item \autoref{prompt:text_cleaning}: Entity label cleaning
\item \autoref{prompt:text-spans}: Text span labeling
\item \autoref{prompt:timechat}: Temporal span labeling
\item \autoref{prompt:spatial-spans}: Spatial span labeling (VLM prompt)
\end{enumerate}

\begin{figure*}
\noindent\fbox{%
    \parbox{\textwidth}{%
\textbf{Entity label cleaning}\\

\footnotesize
{\tt
You are a data cleaning system that takes in natural language descriptions of images and shortens them to labels of visual entities for an object classification model to train on. Keeping as much meaning from the original descriptions as possible, shorten the following passages to one- to three-word image labels. Return your labels in list format.\\

1. <original label 1>\\

2. <original label 2>\\

3. <original label 3>
\\
}
}}

\caption{}
\label{prompt:text_cleaning}
\end{figure*}
\begin{figure*}
\noindent\fbox{%
    \parbox{\textwidth}{%
\textbf{Text span retrieval}\\

\footnotesize
{\tt
You are generating labels for a SQuAD style dataset, a question-answer dataset where the answers are short snippets from one of a collection of multilingual text documents about a {activity}. The label generation process involves two steps: Identifying potential answers to the question, and then comparing against other documents to select the most accurate answer. You are working on step one: You are provided with a question and a text document, and your job is to identify all potential answers in the text.\\

Your answers should only contain text from the original passage. It is good to be as concise as possible - select the smallest passages that still contain the necessary information. The text should generally be a complete phrase, e.g. write “the president” instead of “president”. For additional prepositional phrases, only include these if they add information relevant to the template, e.g. “the president of the United States”.\\

If there are no suitable potential answers in the text document, write N/A. Unless your response is N/A, write your answer in list format, i.e.\\

1. ["Answer 1", "Answer 2", ...]\\
2. N/A\\
3. ["Answer 1"]\\
4. ...\\

TEXT PASSAGE: "<text passage>"\\

QUESTIONS: <template questions>\\

ANSWERS:
}
}}

\caption{}
\label{prompt:text-spans}
\end{figure*}
\begin{figure*}
\noindent\fbox{%
    \parbox{\textwidth}{%
\textbf{Temporal content retrieval}\\

\footnotesize
{\tt
Localize the visual content described by the given textual query Event: <event type>, Query: <template question> in the video, and output the start and end timestamps in seconds. The output format of the predicted timestamp should be like: 'start - end seconds'. A specific example is : 20.8 - 30.0 seconds' .
}
}}

\caption{}
\label{prompt:timechat}
\end{figure*}
\begin{figure*}
\noindent\fbox{%
    \parbox{\textwidth}{%
\textbf{Spatial content retrieval}\\

\footnotesize
{\tt
You are generating candidate answer labels for a visual QA dataset that a human will quality-check. Given the image of a <event type> and the following question, provide any answer candidates you can see in the image in list format and nothing else. If you cannot use the image to answer the question, write N/A.\\

Question: "<template question>"
}
}}

\caption{}
\label{prompt:spatial-spans}
\end{figure*}
\section{\dsetname { } annotation instructions}
\label{appendix:instructions}

\subsection{Introduction}
\label{sub-appendix:instructions:introduction}
When learning about a new current event, we often search for information that helps us answer specific questions about it. For example, if learning about a recent wildfire, we might ask questions like “where was the fire”, “when did the fire take place”, and “who was affected by the fire”. Notably, the type of current event we consider will influence the set of questions we wish to answer— for instance, the sort of questions we care about when learning about a press conference are not the same as those we’d care about when learning about a wildfire.

Internet videos and their corresponding descriptions (e.g. YouTube videos) often contain rich information about current events and are used by many to keep up-to-date with news. In this task, we are interested in annotating exactly how videos and text help us answer event-based questions like those described above. You will be provided with a set of video-text pairs in a language you are an expert in via an annotation interface. Each video-text pair has already been assigned a set of relevant event-based questions called a template. \textbf{Your goal is to identify the text and visual content that would help a person answer these questions in the template.} In the following sections, we will walk you through how to annotate a video and text description to achieve this goal.

\subsection{Task Overview}
\label{sub-appendix:instructions:task-overview}

As mentioned in the introduction, the goal in this task is to identify the text and visual content that would help a person answer the questions in the provided template. Each video-text pair will have a corresponding template (set of relevant questions) that will be used to annotate it. Each template typically has around six questions, or template fields. In the annotation interface, these template fields are referred to with a short name such as “what”, “where”,, or “who-affected”, but each of these corresponds to a full question that can be found in the template guide located at the end of this document. You are recommended to review the template guide often and use it as a reference when you are annotating. 

But, you may be asking: What sort of text and visual content are we annotating, and how do we decide whether or not it “would help” a person answer event-based questions?

We can divide our general task into three subtasks based on the three modalities of data we will be identifying and annotating:

\begin{enumerate}
\item Phrases in text documents
\item Segments of time in videos
\item Bounding boxes in video frames
\end{enumerate}

For each modality, the general question is the same: What [text phrases/video segments/objects in this image] provide information pertaining to each template field? Sometimes, a piece of data will clearly answer a question covered by a template field. For example, in a video of a protest, you will likely see people holding signs and marching - these people clearly provide information about the ``WHO" field of the ``DEMONSTRATION" template (see the template guide at the end of the document).

Other times, a piece of data will provide more implicit information pertaining to a template field that can nevertheless help answer the relevant event-based question. For example, in a news report of a hurricane, the text description may include the phrase ``Hurricane last Monday". While this phrase does not directly give a complete answer to the question ``when did the emergency occur" (corresponding to the ``WHEN" field of the `DISASTER" template), we would still want to label this text phrase as providing information pertaining to the ``WHEN" field.

Depending on the video you are annotating, much of the visual content may be confusing, and it might not always be clear whether or not visual content is salient to a template field or not. We will cover this case in detail in the frame annotation section, but generally, you will want to annotate ambiguous content if you feel there is a nonzero chance that it pertains to a template field. This is generally not as much of an issue with text, which tends to be more direct and unambiguous.

Now we are ready to work through the full annotation pipeline, which is detailed over the next few sections of the guide.

\subsection{Step 0: Review the template and data}
\label{sub-appendix:instructions:step-0}

Before annotating, in addition to reviewing the template guide, you should also review the data by reading the provided text description and watching the video (with sound). Even though we do not explicitly annotate audio in this task, hearing it yourself while reviewing the video is important for obtaining all the context necessary to interpret the data accurately.

\subsection{Step 1: Text annotation}
\label{sub-appendix:instructions:step-1}

After familiarizing yourself with the data, the first annotation step is to identify all phrases in the text document that help answer any of the provided template fields. To annotate text, 

\begin{enumerate}

\item First click on the template field that you want to label a text section with.

\item Then, just highlight the section of text that provides information about that template field. It is good to be as concise as possible - select the smallest passage that still contains all the relevant information associated with that template field. Note: the text should generally be a complete phrase, e.g. annotating “the president” instead of “president”. For additional prepositional phrases, only include these if they add information relevant to the template, e.g. “the president of the United States”.
\end{enumerate}

If two phrases both help answer the same field but describe two different entities, both should be separately highlighted. Similarly, if two sections of text describe the same relevant entity but are separated by other text describing a separate entity, both sections should be highlighted separately. It is likely that you’ll come across a phrase that provides information about multiple template fields. In this case, you should highlight it multiple times (once per relevant template field).

\textbf{For hashtags} (e.g. \#NewYork, \#BTS, etc.): Please include the hashtag symbol (\#) in your spans when they precede a relevant text span.

\subsection{Step 2: Timeline annotation}
\label{sub-appendix:instructions:step-2}

The goal of this step is to identify and record what portions of the video provide information about the template fields. We call these video portions “time segments”, and they can each be defined as a pair of “start” and “end” time points. Specifically, you will identify a segment that contains visual information pertinent to a question, mark the start and end points that this information appears on-screen, and then label it with the template field it provides information for. \uline{Even though you’re shown the audio information in this module, you should not annotate any audio information - in this subtask, we only want to annotate visual data that appears in the video.} To annotate the timeline,

\begin{enumerate}
\item Pause the video where you want to mark the starting point.

\item Click on the question field that matches the visual content.

\item Drag from this starting point on the timeline to the approximate endpoint.

\item To pinpoint the exact endpoint, you can then pause at the precise endpoint, and then re-drag the right side of the segment to this endpoint.
\end{enumerate}

\subsection{Step 3A: Frame selection}
\label{sub-appendix:instructions:step-3a}

For each timeline segment that you created in step 2, you should find a frame within it that clearly displays the salient visual content that you annotated that segment for. If two segments overlap, it is fine to select one frame that satisfies both, as long as the visual content for both segments is clearly pictured in the frame. The frame can be from any point within the segment as long as it is between the identified start and end points.

\subsection{Step 3B: Drawing bounding boxes}
\label{sub-appendix:instructions:step-3b}

After selecting a frame that clearly depicts the relevant visual information of one or more annotated timeline segments, the final step is to draw one bounding box on that frame per timeline segment the frame corresponds to. The box should contain the entire relevant visual entity (e.g. if the visual information is a person, the entire visible portion of the person in that frame should be contained within the drawn bounding box, not just the face, etc.). Similarly, the box should not be larger than what is required to contain the visual content (e.g. if only the person’s face is visible, the box should not try to contain what portion of the person’s body isn’t visible in the frame). You should draw one box for each time segment the frame was chosen for, so after drawing each bounding box, there should be one (or, rarely, multiple) box for each time segment you annotated. To draw a bounding box, 

\begin{enumerate}
\item First, go to your selected frame and drag over the video player to draw a box bounding a relevant entity.

\item Letting go of the cursor will produce a gray square. Click inside it to reveal bounding box-specific options. First, if you have the frame-level timeline toggled, you will see a diamond with a line attached to it here:

\item You will want to remove this line for each box you make, which can be done by clicking on the “toggle interpolation” button here while the corresponding box is selected.

\item Then, click on the question field below the frame timeline that corresponds to the bounding box.

\end{enumerate}

\subsection{Step 3C: Annotating bounding box details}
\label{sub-appendix:instructions:step-3c}

After drawing these boxes, you will be asked to answer three questions about each bounded entity:

\begin{enumerate}
\item ENTITY DESCRIPTION: Give a short description of the bounded content, ideally one or two words. When you have typed the description, click the “Add” button to save it. This is the one annotation in the task that must be manually saved. If you feel that more than one description is necessary (for example, if the bounded entity is a doctor who has a visible nametag, you ideally will add “doctor” as well as their name) then after “Add”ing the first description, type in a second description and click “Add” again.

\item IS THIS TEXT: Select “Yes” if your bounding box is an image of text, e.g. the words on a street sign. Here are examples of text that would be labeled as such in a video:

\item CONFIDENCE SCORE: Rate your confidence that the bounding box helps answer the question field you selected. More information on confidence scores can be found in the next section of the guide.

\item ADDITIONAL NOTES: This section is optional and can be used to add any additional notes that you think are important to log about the bounded entity. Feel free to leave this blank most of the time, if not always.
\end{enumerate}

\subsection{Confidence scores}
\label{sub-appendix:instructions:confidence}

Visual data is difficult to annotate in that there is often uncertainty about what a visual entity is and what its relationship to the event is. In this task, we consider this uncertainty in terms of saliency. Given any visual entity in a video, there is an underlying probability that the entity answers a given template question (who,what,where…).

This is the purpose of the confidence scoring for individual bounding boxes. Given a bounding box and the template question it answers, you will rate how confident you are that the bounded entity does actually help to answer the template field question on a scale from 1 to 5.

This scale roughly aligns with the probability that the bounded entity is salient. Giving a boxed entity a confidence score of 2 means that you believe there is roughly a 40\% chance that the entity helps answer the template field question. This system can be considered in terms of gambling - if you had \$5, and were aiming to earn/keep as much money as possible, how many dollars would you bet that the entity answers the template question? If you were 100\% certain, you would bet \$5 (and mark your confidence as 5). If you thought that it was likely, but not certain, you might bet \$4 (and give a 4/5 rating). If you thought there was a low chance, but was still possible, you might bet \$1 (1/5).

Because we are interested in collecting confidence scores, this means that you should annotate visual content that you are not certain about in the videos. Anything that looks like it may be salient should be annotated and rated accordingly. However, if you are pretty certain that something is not salient (less than 20\% sure), you do not need to annotate it. This is why there is no “0” on the confidence scale.

These ratings are subjective, and so your confidence may be different than another annotator’s. As long as you are rating your own confidence that the bounded content helps answer the corresponding template question, your annotation is correct.

\subsection{Confidence score examples}
\label{sub-appendix:instructions:confidence-exs}

Below, we have included some examples of confidence ratings. Note that these confidence scores are based on still images, whereas you will have the full context of the video and text, and so there is a slight difference in how you may make your certainty decisions compared to these examples.

\begin{figure}
\includegraphics[width=\linewidth]{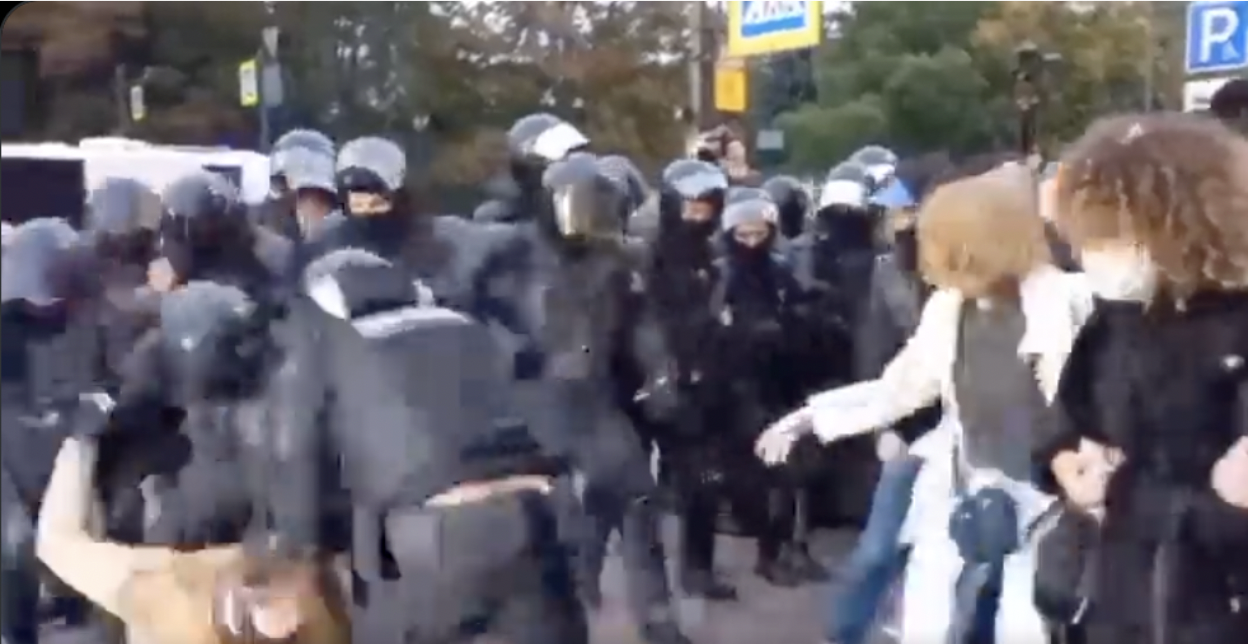}
\caption{An example video frame used to explain confidence judgements in the annotation instructions manual.}
\label{fig:confidence-example}
\end{figure}

Pretend I am annotating a video of a protest. I know the video is of a protest, but I am only provided with this single frame from the video (Figure~\ref{fig:confidence-example}).

I am confident that the people dressed in black are law enforcement officers present at the protest. Therefore, I can box them, tag them with the “LAW-ENFORCEMENT” field, and mark my confidence as 5/5.

However, I am less certain about whether the figure in the bottom left is a protester (or even a person?). I think it is fairly likely, since it looks like the shape of a person and the law enforcement officers are surrounding them, and so I’d tag them as a “WHO” and would probably rate my confidence as 3/5. Possibly 4/5, but definitely not 5/5 or 1/5.

Lastly, my best guess is that the people on the right of the frame are protesters, since they are near the other person I think is a protester being arrested. However, I am even less certain about these people. Perhaps the demonstration was being conducted by a single person, and these were just onlookers. Therefore, I’d tag these people as “WHO” with a confidence score of 2/5.

As you can see, all of these decisions are made based on personal knowledge and reasoning, and so they may vary slightly from person to person. This variance is perfectly okay.

\subsection{Frequently asked questions}
\label{sub-appendix:instructions:faqs}

\begin{itemize}
\item \textbf{What language should I use for labeling bounding boxes?} If the bounded entity is a piece of actual text, please transcribe it in the original language. Otherwise, please write a brief description in English. When transcribing, you do not need to also translate the text.
\item \textbf{If I want to draw bounding boxes for a group of people, should I draw one large box or individual boxes?} For groups of objects in general, unless they (a) answer different template field questions, (b) they have distinct identities that are clear from the video, and/or (c) they are significantly separated by other entities in the frame, please draw one bounding box for all of them. Otherwise, it might be preferable to draw individual bounding boxes, unless there is a significantly large number of objects present (e.g. more than 6).
\item \textbf{When writing multiple labels for a bounding box, does the order that they are entered matter?} No.
\item \textbf{What do I do if a single bounding box applies to multiple questions/template fields?} In this case, you should make multiple bounding boxes over the same area and tag each with a different field. This also applies for text and time segments.
\item \textbf{Some text implies that an entity that would answer a template field exists, but does not directly state it. Should I annotate this text and label it with that field?} Yes. This also applies to other visual data that implies things that would answer template fields.
\item \textbf{What if there isn’t any text for a video-text pair?} In that case, please leave the text annotation field blank and move on to the next section.
\end{itemize}
\section{Ontology}
\label{appendix:ontologies}

Here we provide the \dsetname{} ontology in \autoref{tab:multivent-ontology}.

\begin{table*}[t!]
    \small
    \centering
    \caption{\dsetname { }ontology, as written in the annotation instructions manual.}
    \begin{tabular}{lp{7cm}p{5cm}}
    \toprule
        \textbf{Field Name} & \textbf{Field Description} & \textbf{Examples} \\
        \midrule
        
        \multicolumn{3}{c}{\textbf{Disasters}}\\
    \midrule
        \textbf{What} & What emergency/disaster is occurring/has occurred? & fire, earthquake, hurricane \\
        \textbf{Where} & Where is the emergency/disaster occurring? & Australia, USA, movie theater \\
        \textbf{When} & When did the emergency/disaster occur? & morning, last week, 2020 \\
        \textbf{Outcome-Occurred} & What was the outcome of the emergency/disaster? & burned trees, collapsed buildings \\
        \textbf{Who-Affected} & Who was affected by the emergency/disaster? & people huddled together, animals running for cover, person interviewed \\
        \textbf{Emergency-Response} & Who is responding to assist those affected by the emergency/disaster? & firefighters, doctors attending to wounded, politicians at the scene \\
\midrule
        \multicolumn{3}{c}{\textbf{Elections}}\\
    \midrule
        \textbf{Body} & What place/population does the election pertain to? & Taiwan, US soccer federation \\
        \textbf{Role} & What is the position of the elected leader? & President, senator \\
        \textbf{New-Leader} & Who is the elected leader? & Barack Obama \\
        \textbf{Old-Leader} & Who is the elected leader replacing? & George W. Bush \\
        \textbf{Selector} & Who is electing the new leader? & Ukrainian citizens \\
        \textbf{When} & When does the election take place? & Morning, last week, 2020 \\
        \textbf{Where} & Where does the voting take place? & US, polling center, online \\
        \textbf{Candidate} & Who is running as a candidate in the election? & Joe Biden, Ted Cruz \\
\midrule
        \multicolumn{3}{c}{\textbf{Political Developments}}\\
    \midrule
        \textbf{Development} & What is the development? & Sanctions, leader death \\
        \textbf{Cause} & What has caused the development? & Trade war, protests \\
        \textbf{Effect} & What is the effect of the development? & Economy collapse, new leader \\
        \textbf{Agent(s)} & Who participated in causing the development? & Biden, ambassadors, citizens \\
        \textbf{Affected-Place} & Where does the development affect? & Russia, international waters \\
        \textbf{Time} & When does the development take place? & Morning, last week, 2020 \\
\midrule
        \multicolumn{3}{c}{\textbf{Social Events}}\\
    \midrule
        \textbf{Attendees} & Who has come to participate in the event? & Children, people, senators \\
        \textbf{Social-Event} & What is the event? & Coachella, Ramadan \\
        \textbf{Place} & Where does the event take place? & Taiwan, a hotel \\
        \textbf{Time} & When is the event taking place? & Morning, last week, 2020 \\
        \textbf{Event-Artifact} & What physical entities have been constructed or are presented on behalf of the event? & Art exhibit, information stand \\
        \textbf{Performer/Staff} & Who is working or presenting on behalf of the event? & Rihanna, speaker, staff \\
     \midrule    
        \multicolumn{3}{c}{\textbf{Sports}}\\
    \midrule
        \textbf{Competition} & What competition is it? & World Cup 2022 \\
        \textbf{Participants} & Who is participating in the competition? & Players, Brazil, Yankees \\
        \textbf{Place} & Where is the competition taking place? & Qatar, stadium, Lusail \\
        \textbf{Time} & When does the competition take place? & Morning, last week, 2020 \\
        \textbf{Result} & What scores/results occur during the competition? & 4-5, won, tie \\
        \textbf{Spectators} & Who comes to watch the competition? & People, Yankees fans \\
        \textbf{Sport} & What sports are played in the competition? & Soccer, pole vault, marathon \\
\midrule
        \multicolumn{3}{c}{\textbf{Discoveries/Launches}}\\
    \midrule
        \textbf{Developer} & Who has developed/discovered the concept? & Amazon, an archaeologist \\
        \textbf{What} & What is being launched/discovered? & Black hole image, iPhone 92 \\
        \textbf{Place} & Where was the concept developed/discovered? & Cupertino, US, a volcano \\
        \textbf{Time} & When was the concept developed/discovered? & Morning, last week, 2020 \\
        \textbf{Use} & What is the use case of this concept? & Self-driving car, a physics theory \\
        \textbf{Presentation-Location} & Where was the concept launched/presented? & A conference, Thailand \\
\midrule
        \multicolumn{3}{c}{\textbf{Demonstrations}}\\
    \midrule
        \textbf{Who} & Who are the protesters? & Teenagers, people, veterans \\
        \textbf{Where} & Where is the protest occurring? & Moscow, the street, the white house \\
        \textbf{When} & When did the protest occur? & Lasted for 12 minutes, March 2020 \\
        \textbf{Organization} & What organizations are involved with the protest? & Truckers association, ANTIFA \\
        \textbf{Issue} & What issue does the protest concern? & Taxes, war, climate change \\
        \textbf{Law-Enforcement} & What law enforcement was involved at the protest? & Security guards, OMON \\
         
        \bottomrule
    \end{tabular}
    \label{tab:multivent-ontology}
\end{table*}

\section{LLM Samples}

We provide (a) a collection of sample outputs from the text labeling task in \autoref{prompt:sample_outs} and (b) the few-shot samples provided for ICL prompting in \autoref{sec:experiments}:

\begin{enumerate}
\item \autoref{prompt:emergency}: Disasters
\item \autoref{prompt:election}: Elections
\item \autoref{prompt:political}: Political developments
\item \autoref{prompt:social}: Social events
\item \autoref{prompt:sports}: Sports
\item \autoref{prompt:discovery}: Discoveries/launches
\item \autoref{prompt:demonstrations}: Demonstrations
\end{enumerate}

\begin{figure*}
\noindent\fbox{%
    \parbox{\textwidth}{%
\textbf{Input prompt}\\

\footnotesize
{\tt
 QUESTIONS: \\
1. "What emergency/disaster is occurring/has occurred?"\\
2. "Where is the emergency/disaster occurring?"\\
3. "When did the emergency/disaster occur?"\\
4. "Who was affected by the emergency/disaster?"\\
5. "Who is responding to assist those affected by the emergency/disaster?"\\

TEXT PASSAGE: "The opening of the floodgates of the Ross River dam in Queensland created what Australia's weather bureau called "dangerous and high velocity flows". With the dam at more than 300 percent capacity, engineers said they had no option.\\

Downstream homes in the coastal city of Townsville are flooded and more than a thousand people have moved to safety and been evacuated from their homes.\\
In some parts of Townsville, salt-water crocodiles displaced by the floodwaters have been spotted, adding extra risk and danger. \\

Al Jazeera\'s Bernard Smith reports.\\

- Subscribe to our channel: http://aje.io/AJSubscribe \\
- Follow us on Twitter: https://twitter.com/AJEnglish \\
- Find us on Facebook: https://www.facebook.com/aljazeera\\ 
- Check our website: https://www.aljazeera.com/\\

\#AlJazeeraEnglish \#Australia \#Floods" \\
 
 ANSWERS:
}
}}
\vspace{5mm}

\noindent\fbox{%
    \parbox{\textwidth}{%
\textbf{GPT-3.5}\\

\footnotesize
{\tt
1. ["floods", "dangerous and high velocity flows"]\\
2. ["coastal city of Townsville"]\\
3. N/A\\
4. ["more than a thousand people"]\\
5. ["engineers", "Al Jazeera's Bernard Smith"]
}
}}
\vspace{5mm}

\noindent\fbox{%
    \parbox{\textwidth}{%
\textbf{GPT-4o}\\

\footnotesize
{\tt
1. ["The opening of the floodgates of the Ross River dam", "floodwaters"]\\
2. ["Queensland", "the coastal city of Townsville", "Townsville"]\\
3. N/A\\
4. ["Downstream homes in the coastal city of Townsville", "more than a thousand people", "people have moved to safety and been evacuated from their homes"]\\
5. N/A
}
}}
\vspace{5mm}

\noindent\fbox{%
    \parbox{\textwidth}{%
\textbf{GPT-4}\\

\footnotesize
{\tt
1. ["floods", "dangerous and high velocity flows"]\\
2. ["coastal city of Townsville"]\\
3. N/A\\
4. ["more than a thousand people"]\\
5. ["engineers", "Al Jazeera's Bernard Smith"]
}
}}

\caption{}
\label{prompt:sample_outs}
\end{figure*}
\begin{figure*}
\noindent\fbox{%
    \parbox{\textwidth}{%
\textbf{Few-shot samples: DISASTERS}\\

\footnotesize
{\tt
QUESTIONS: \\
1. "What emergency/disaster is occurring/has occurred?"\\
2. "Where is the emergency/disaster occurring?"\\
3. "When did the emergency/disaster occur?"\\
4. "Who was affected by the emergency/disaster?"\\
5. "Who is responding to assist those affected by the emergency/disaster?"\\

TEXT PASSAGE: "Europe's most active Campi Flegrei volcano experienced its largest seismic swarm in 40 years, with 150 earthquakes shaking southern Italy on the evening of May 20. Engineers are responding to infrastructural damage in the area." \\

Join the Community https://bit.ly/godmembership\\

\#volcano \#naples \#supervolcano" \\
 
ANSWERS:\\
1. ["its largest seismic swarm", "150 earthquakes", "\#volcano", "\#supervolcano"]\\
2. ["Europe's most active Campi Flegrei volcano", "southern Italy", "\#naples"]\\
3. ["40 years", "the evening of May 20"]\\
4. N/A\\
5. ["engineers"]
}
}}

\caption{Prompt adapted from text in https://www.youtube.com/watch?v=svsmOvbXAPQ}
\label{prompt:emergency}
\end{figure*}

\begin{figure*}
\noindent\fbox{%
    \parbox{\textwidth}{%
\textbf{Few-shot samples: ELECTIONS}\\

\footnotesize
{\tt
QUESTIONS: \\
1. "What place/population does the election pertain to?" \\
2. "What is the position of the elected leader?" \\
3. "Who is the elected leader?" \\
4. "Who is the elected leader replacing?" \\
5. "Who is electing the new leader?" \\
6. "When does the election take place?" \\
7. "Where does the voting take place?" \\
8. "Who is running as a candidate in the election?" \\

TEXT PASSAGE: "Climate scientist and former Mexico City mayor Claudia Sheinbaum is projected to be Mexico's presidential election winner, making her the country's first female president. Election day was on June 2, 2024, and her main opponent was former Senator Xóchitl Gálvez. Sheinbaum has long been an ally of the incumbent president, Andrés Manuel López Obrador.\\

CBS Mornings airs weekdays at 7 a.m. on CBS and stream it at 8 a.m. ET on the CBS News app." \\
 
ANSWERS:\\
1. ["Mexico", "the country", "Mexico City"]\\
2. ["president"]\\
3. ["Climate scientist", "former Mexico City mayor Claudia Sheinbaum", "Sheinbaum"]\\
4. ["Andrés Manuel López Obrador"]\\
5. N/A\\
6. ["June 2, 2024"]\\
7. N/A\\
8. ["Climate scientist", "former Mexico City mayor Claudia Sheinbaum", "former Senator Xóchitl Gálvez", "Sheinbaum"]
}
}}

\caption{Prompt adapted from text in https://sg.news.yahoo.com/mexico-elects-1st-female-president-following-deadliest-election-campaign-in-countrys-modern-history-heres-what-to-know-184442748.html}
\label{prompt:election}
\end{figure*}

\begin{figure*}
\noindent\fbox{%
    \parbox{.98\textwidth}{%
\textbf{Few-shot samples: POLITICAL DEVELOPMENT}\\

\footnotesize
{\tt
QUESTIONS: \\
1. "What is the development?" \\
2. "What has caused the development?" \\
3. "What is the effect of the development?" \\
4. "Who participated in causing the development?" \\
5. "Where does the development affect?" \\
6. "When does the development take place?" \\

TEXT PASSAGE: "Thaksin Shinawatra, the former prime minister of Thailand and still the most influential figure in the ruling Pheu Thai Party, has been indicted on lèse-majesté charges, revealing the latest twist in the country’s fragile political landscape." \\
 
ANSWERS:\\
1. ["indicted on lèse-majesté charges"]\\
2. ["the country's fragile political landscape"]\\
3. ["the latest twist in the country's fragile political landscape"]\\
4. ["Thaksin Shinawatra", "the former prime minister of Thailand", "the most influential figure in the ruling Pheu Thai Party"]\\
5. ["Thailand", "the country"]\\
6. N/A
}
}}

\caption{Prompt adapted from text in https://www.cfr.org/blog/has-king-thailand-split-thaksin}
\label{prompt:political}
\end{figure*}

\begin{figure*}
\noindent\fbox{%
    \parbox{.98\textwidth}{%
\textbf{Few-shot samples: SOCIAL EVENTS}\\

\footnotesize
{\tt
QUESTIONS: \\
1. "Who has come to participate in the event?" \\
2. "What is the event?" \\
3. "Where does the event take place?" \\
4. "When is the event taking place?" \\
5. "What physical entities have been constructed or are presented on behalf of the event?" \\
6. "Who is working or presenting on behalf of the event?" \\

TEXT PASSAGE: "Watch as stars walk the 2024 Met Gala carpet and leave The Mark Hotel in New York. Notable names include Zendaya, Doja Cat, and Anna Wintour.\\

\#metgala \#metgala2024" \\
 
ANSWERS:\\
1. ["stars", "Zendaya", "Doja Cat", "Anna Wintour"]\\
2. ["the 2024 Met Gala"]\\
3. ["The Mark Hotel", "New York"]\\
4. ["2024"]\\
5. ["the 2024 Met Gala carpet"]\\
6. N/A
}
}}

\caption{Prompt adapted from text in https://www.youtube.com/watch?v=JIGkDU7Z39c}
\label{prompt:social}
\end{figure*}

\begin{figure*}
\noindent\fbox{%
    \parbox{.98\textwidth}{%
\textbf{Few-shot samples: SPORTS}\\

\footnotesize
{\tt
QUESTIONS: \\
1. "What competition is it? \\
2. "Who is participating in the competition?" \\
3. "Where is the competition taking place?" \\
4. "When does the competition take place?" \\
5. "What scores/results occur during the competition?" \\
6. "Who comes to watch the competition?" \\
7. "What sports are played in the competition?" \\

TEXT PASSAGE: "The Orioles led MLB with 41 home runs in the month of April. Reigning Rookie of the Year Gunnar Henderson led the team with nine homers, with outfielder Colton Cowser following close behind with six homers.\\

Subscribe to the YT Channel: https://bit.ly/2SYEQEV\\

\#BaltimoreOrioles \#Birdland" \\
 
ANSWERS:\\
1. ["MLB"]\\
2. ["The Orioles", "Gunnar Henderson", "Colton Cowser", "\#BaltimoreOrioles"]\\
3. N/A\\
4. ["the month of April"]\\
5. ["41 home runs", "nine homers", "six homers"]\\
6. N/A\\
7. ["MLB"]
}
}}

\caption{Prompt adapted from text in https://www.youtube.com/watch?v=672-6MK6Xu8}
\label{prompt:sports}
\end{figure*}

\begin{figure*}
\noindent\fbox{%
    \parbox{.98\textwidth}{%
\textbf{Few-shot samples: DISCOVERY/LAUNCH}\\

\footnotesize
{\tt
QUESTIONS: \\
1. "Who has developed/discovered the concept?" \\
2. "What is being launched/discovered?" \\
3. "Where was the concept developed/discovered?" \\
4. "When was the concept developed/discovered?" \\
5. "What is the use case of this concept?" \\
6. "Where was the concept launched/presented?" \\

TEXT PASSAGE: "A team of researchers from South Korea created a lightweight structure to improve the energy density of lithium-ion batteries, which will enable them to remain stable for longer, Tech Xplore reported. Their results were published in the journal Advanced Science in April." \\
 
ANSWERS:\\
1. ["A team of researchers from South Korea"]\\
2. ["a lightweight structure"]\\
3. ["South Korea"]\\
4. ["April"]\\
5. ["to improve the energy density of lithium-ion batteries", "enable them to remain stable for longer"]\\
6. ["the journal Advanced Science"]
}
}}

\caption{Prompt adapted from text in https://www.yahoo.com/tech/ev-battery-researchers-unveil-lightweight-003000555.html}
\label{prompt:discovery}
\end{figure*}

\begin{figure*}
\noindent\fbox{%
    \parbox{.98\textwidth}{%
\textbf{Few-shot samples: DEMONSTRATIONS}\\

\footnotesize
{\tt
QUESTIONS: \\
1. "Who are the protesters?" \\
2. "Where is the protest occurring?" \\
3. "When did the protest occur?" \\
4. "What organizations are involved with the protest?" \\
5. "What issue does the protest concern?" \\
6. "What law enforcement was involved at the protest?" \\

TEXT PASSAGE: "BRUSSELS, June 4 (Reuters) - Farmers and others from European agricultural groups in Brussels on Tuesday protested against the European Union's environmental policies, but the action was shunned by mainstream farming groups who said it did not reflect their members' concerns.\\

A few days before the European Parliament election, farmers from the Netherlands travelled to Brussels to protest against EU green policies that organisers said undermine the competitiveness of European farmers." \\
 
ANSWERS:\\
1. ["Farmers", "others", "farmers from the Netherlands", "European farmers"]\\
2. ["Brussels"]\\
3. ["June 4", "Tuesday", "A few days before the European Parliament election"]\\
4. ["European agricultural groups", "mainstream farming groups"]\\
5. ["the European Union's environmental policies", "EU green policies"]\\
6. N/A
}
}}

\caption{Prompt adapted from text in https://www.reuters.com/world/europe/main-farming-groups-shun-brussels-protest-against-eu-green-policies-2024-06-04/}
\label{prompt:demonstrations}
\end{figure*}
\end{document}